\documentclass[letterpaper, 10 pt, conference]{ieeeconf}  

\IEEEoverridecommandlockouts                              

\usepackage{multirow}
\usepackage{multicol}

\usepackage{hyperref}
\usepackage{graphicx}
\usepackage{color}
\usepackage{adjustbox}
\usepackage{amsmath}

\usepackage{algorithmic}
\usepackage{graphicx}
\usepackage{textcomp}
\usepackage{xcolor}
\usepackage{caption}
\usepackage{subcaption}
\usepackage{makecell}
\usepackage{float}
\usepackage{makecell}
\usepackage{booktabs}
\usepackage{soul}
\usepackage{booktabs}

\graphicspath{{images/}}

\title{\LARGE \bf SocNavGym: A Reinforcement Learning Gym for Social Navigation}

\author{Aditya Kapoor$^{*1}$, Sushant Swamy$^{*2}$, Pilar Bachiller$^{3}$ and Luis J. Manso$^{4}$
\thanks{$^{1}$Aditya Kapoor is with Tata Consultancy Services, Research \& Innovation, India.
        {\tt\small aditya.kapoor1@tcs.com}}%
\thanks{$^{2}$Sushant Swamy is with Birla Institute of Technology and Science, Goa, India.
        {\tt\small f20191031@goa.bits-pilani.ac.in}}%
\thanks{$^{3}$Pilar Bachiller is with Computer and Communication Technology Department, Universidad de Extremadura, Spain.
        {\tt\small pilarb@unex.es}}%
\thanks{$^{4}$Luis J. Manso is with Autonomous Robotics and Perception Laboratory, Computer Science Department, Aston University, UK.
        {\tt\small l.manso@aston.ac.uk}}%
\thanks{$^{*}$ is for equal contribution.}
}

\begin{document}

\maketitle
\thispagestyle{empty}
\pagestyle{empty}

\begin{abstract}

It is essential for autonomous robots to be socially compliant while navigating in human-populated environments.
Machine Learning and, especially, Deep Reinforcement Learning have recently gained considerable traction in the field of Social Navigation.
This can be partially attributed to the resulting policies not being bound by human limitations in terms of code complexity or the number of variables that are handled.
Unfortunately, the lack of safety guarantees and the large data requirements by DRL algorithms make learning in the real world unfeasible.
To bridge this gap, simulation environments are frequently used.
We propose SocNavGym, an advanced simulation environment for social navigation that can generate a wide variety of social navigation scenarios and facilitates the development of intelligent social agents.
SocNavGym is lightweight, fast, easy to use, and can be effortlessly configured to generate different types of social navigation scenarios.
It can also be configured to work with different hand-crafted and data-driven social reward signals and to yield a variety of evaluation metrics to benchmark agents' performance.
Further, we also provide a case study where a Dueling-DQN agent is trained to learn social-navigation policies using SocNavGym.
The results provide evidence that SocNavGym can be used to train an agent from scratch to navigate in simple as well as complex social scenarios.
Our experiments also show that the agents trained using the data-driven reward function display more advanced social compliance in comparison to the heuristic-based reward function.

\end{abstract}

\section{Introduction}\label{intro}
Social compliance is key to deploying robots, not only for pedestrians' comfort but also to enhance efficiency.
The sheer amount of research in the field is evidence of its importance~\cite{mavrogiannis2021core}.
Until recently, the vast majority of Social Navigation (SN) algorithms were hand-crafted and coded by humans.
One of the limitations of this approach is its poor scalability in terms of the number of variables considered, which is partially because the equations involved become complex to understand, implement and perhaps most importantly, debug.
Other factors that make incorporating other variables challenging include the lack of a principled theory of pedestrians' comfort, and the fact that the state of the world is only partially observable (\textit{e.g.}, pedestrians' hidden goals, intents, mood).
Because of these limitations, the community is experiencing a strong trend toward data-driven methods that can account for more variables.
The most common SN approach making use of learning involves using hand-crafted algorithms in conjunction with learning-based human trajectory predictors.
Examples of these works are \cite{liu_traj_forecast_2022,chen_traj_forecast_2019,alahi_traj_forecast_2016,gupta_traj_forecast_2018,yingfan_traj_forecast_2019,eiffert_traj_forecast_2020}, where actions are planned using search trees and other heuristic-based algorithms relying on estimated human states~\cite{puterman_markov_2005}. Other learning-based works include social compliance estimators, such as the Social Navigation Graph Neural Network (SNGNN)~\cite{bachiller2022graph}.
A third family of promising social navigation algorithms that has gained recent popularity, are those using Deep Reinforcement Learning (DRL) and Learning from Demonstrations (LfD) \cite{CrowdNav,chen_social_nav_2017,zhou_social_nav_2021,qin_dil_2021,vecente_deepsocnav_2021,yigit_dim_2022}.
\par
DRL systems have three especially critical components: 1)~a DRL algorithm, 2)~a neural architecture, and 3)~an environment providing observations and rewards.
While research in DRL fundamentally focuses on the first two aspects, a limitation in the current literature is that none of the reward functions used in the existing gym environments has been designed to reflect overall users' opinions about the contextual social compliance of the agent.
The ones in use are more suitable to solve multi-agent collision avoidance rather than social navigation, which is a different, relatively easier problem.
To the best of our knowledge, the reward functions used in the DRL-based SN literature are all piece-wise defined functions where the piece considering the social compliance of the agent is a simple linear equation or a similar profile function that only consider pedestrians' positions.
We strongly argue that, to learn socially-compliant policies via DRL, the reward function should be carefully designed to consider users' opinions.
\par

Inverse Reinforcement Learning (IRL) has shown promising results in social navigation~\cite{IRL1,IRL2,IRL4}.
If trained with adequate data, a learned reward function can help tackle these issues, as a DRL architecture could be trained to optimise for it, enabling socially-compliant policy learning.
The resulting behaviour could theoretically be as principled and effective as such a learned reward. 
Our hypothesis in this vein is that using a learned model for comfort as a reward function will lead to better social compliance than other simplistic or hand-crafted rewards.
Therefore, our gym environment -SocNavGym- integrates SNGNN-v2~\cite{SNGNN} in its reward function, the only model that we have found in the literature providing step-wise social compliance scores.
\par

The desiderata established for SocNavGym's design is as follows:
\begin{enumerate}
    \item be lightweight, fast and easy to use and configure, with as few software dependencies as possible;
    \item simulate a rich variety of realistic social scenarios so that agents can be used in as many real-life settings as possible;
    \item be easy to modify and extend;
    \item provide a high-quality social reward to ensure that near-optimal policies meet real pedestrians' expectations; and
    \item provide established evaluation metrics to benchmark agents' performance.
\end{enumerate}
To the best of our knowledge, no existing navigation gym satisfies all the aforementioned requirements (see Section~\ref{relatedworks} for details).

\par
SocNavGym is a 2D simulation environment built by extending the OpenAI's MultiAgent Particle Environment (MPE)~\cite{mordatch2017emergence}.
We take inspiration from the existing work in the literature and incorporate some of the most relevant features into SocNavGym.
Examples of additional features include static obstacle-like entities, static and dynamic agent formations, and interactions between different entities (\textit{e.g.}, human-looking-at-laptop, human-talking-with-human).
We also integrate SNGNN-v2~\cite{SNGNN}, a data-driven reward function trained on a users' opinions dataset~\cite{IndoorDataset1,SONATA}. 
Further, we train a DuelingDQN~\cite{DuelingDQN} model to show that our simulation environment can be used to train DRL algorithms in a variety of rich social situations.

\section{Related work}
\label{relatedworks}
Many high-fidelity simulators have been developed to collect data and train models for robotic applications \textit{e.g.}, CARLA~\cite{CARLA}, iGibson~\cite{iGibson_2021}.
Crowd simulation platforms like Nomad~\cite{NOMAD_2014} and Menge~\cite{Menge_2016} are designed to model individuals and crowds.
On the contrary, SEAN~\cite{SEAN2.0_2022} and SocialGym~\cite{SOCIALGYM} are designed to train and evaluate social agents but their reliance on Unity~\cite{Unity} and Robot Operating System (ROS), might not be ideal for many DRL researchers due to additional complexity and computational resources required.
Although these provide good physics simulations, they require a considerable amount of computing power that does not seem justifiable for this specific application.
Also, to focus on decision-making for social navigation, one can abstract away the perception task that further weakens the need for high-fidelity simulators.
SocNavGym and other state-of-the-art SN gyms (\textit{e.g.}, SARL~\cite{CrowdNav}, DSRNN~\cite{chen_social_nav_2017}), use low-fidelity simulators but, unlike SocNavGym, others fail to recreate realistic complex social settings.
In SARL~\cite{CrowdNav}, scenarios are composed of humans forming a circle moving in the diametrically opposite direction using ORCA policy in an otherwise empty space.
In DSRNN~\cite{chen_social_nav_2017}, static and dynamic groups are also present, but they still greatly differ from typical human-populated environments.
SocialGym~\cite{SOCIALGYM} goes a step further to introduce room-like scenarios with corridors and more realistic pedestrian movement.
However, its dependence on ROS~\cite{quigley2009ros} creates additional overhead, and its setup and installation process is not as easy as desired. Further, bringing about any change in the environment requires the user to know the ROS framework which is not ideal.
Additionally, none of these environments provides a principled social reward function that factors in human opinion, which makes agents trained with them, treat humans as dynamic obstacles rather than social beings.

\par
Another gap found in simulation environments is the lack of support for common interactions, such as those among humans and with other inanimate entities.
Such interactions are important while training agents for realistic scenarios since the level of disturbance that an agent can cause depends on these interactions.
For example, an agent could cause considerable disturbance to two people talking (human-human interaction) even if the agent does not enter the personal space of the individuals involved.
These interactions should be taken into consideration, regardless of them being directly observable by the agents or expected to be inferred (SocNavGym supports both options).

\par
With the exception of SocialGym~\cite{SOCIALGYM} and SEAN~\cite{SEAN2.0_2022}, none of the reviewed SN gyms provide sufficient evaluation metrics (\textit{e.g.}, time taken to reach the goal, minimum distance to human, discomfort level) to benchmark social performance.
Even though these gyms provide a subset of the evaluation metrics, they do not provide any metric or reward function accounting for efficiency and human comfort.


\section{Gym design decisions and features}
\begin{figure*}[t]
  \centering
  \subcaptionbox{\centering Before Crowd Dispersal}[.2\linewidth][c]{%
    \includegraphics[width=\linewidth]{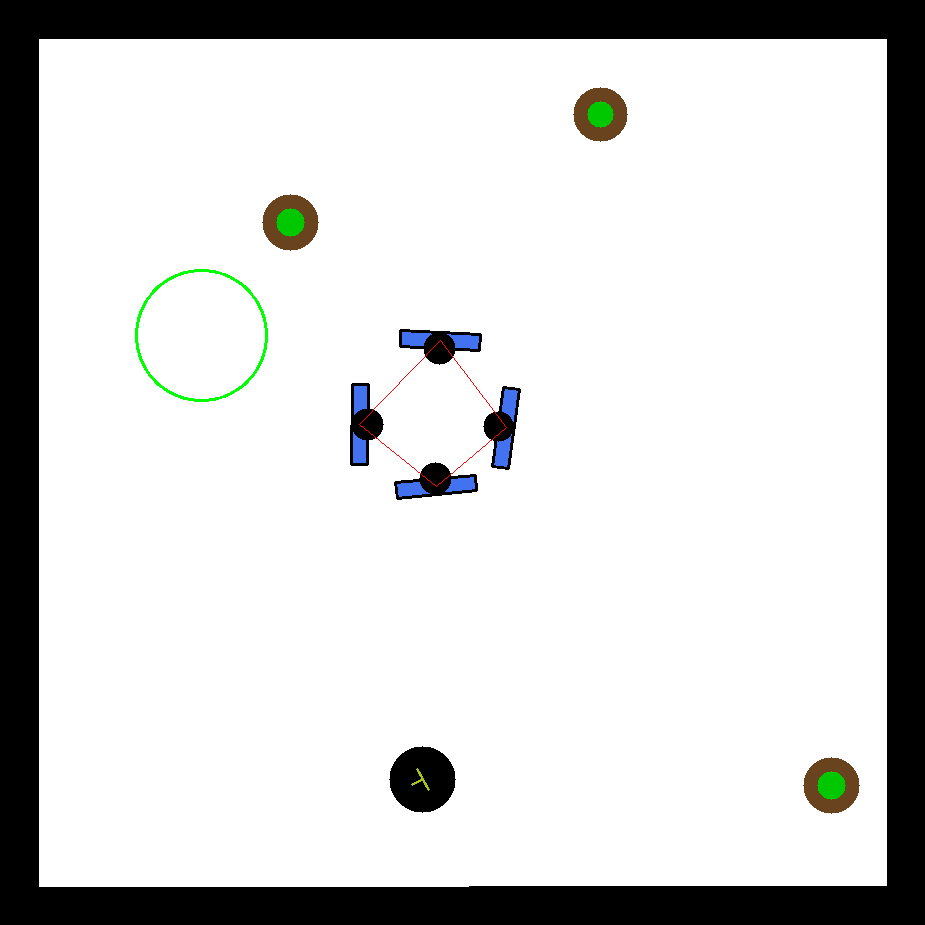}}
  \subcaptionbox{\centering Before Crowd Formation}[.2\linewidth][c]{%
    \includegraphics[width=\linewidth]{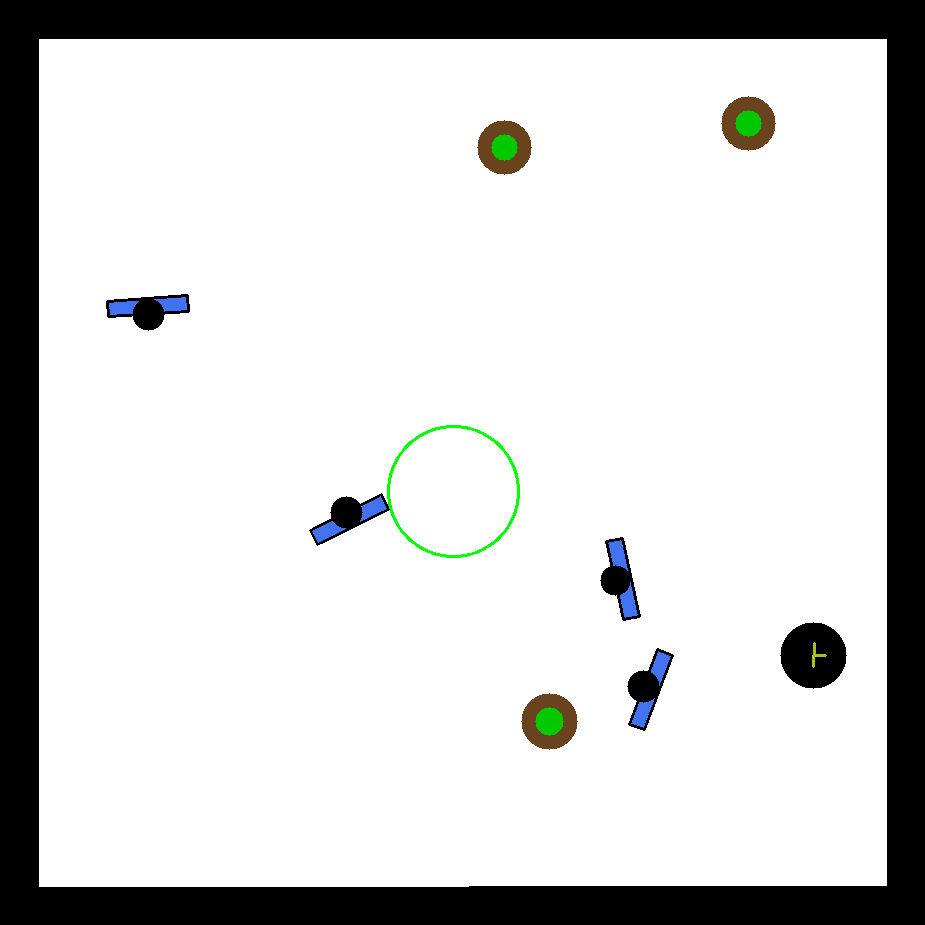}}
  \subcaptionbox{\centering Before Human-Laptop Interaction Dispersal}[.2\linewidth][c]{%
    \includegraphics[width=\linewidth]{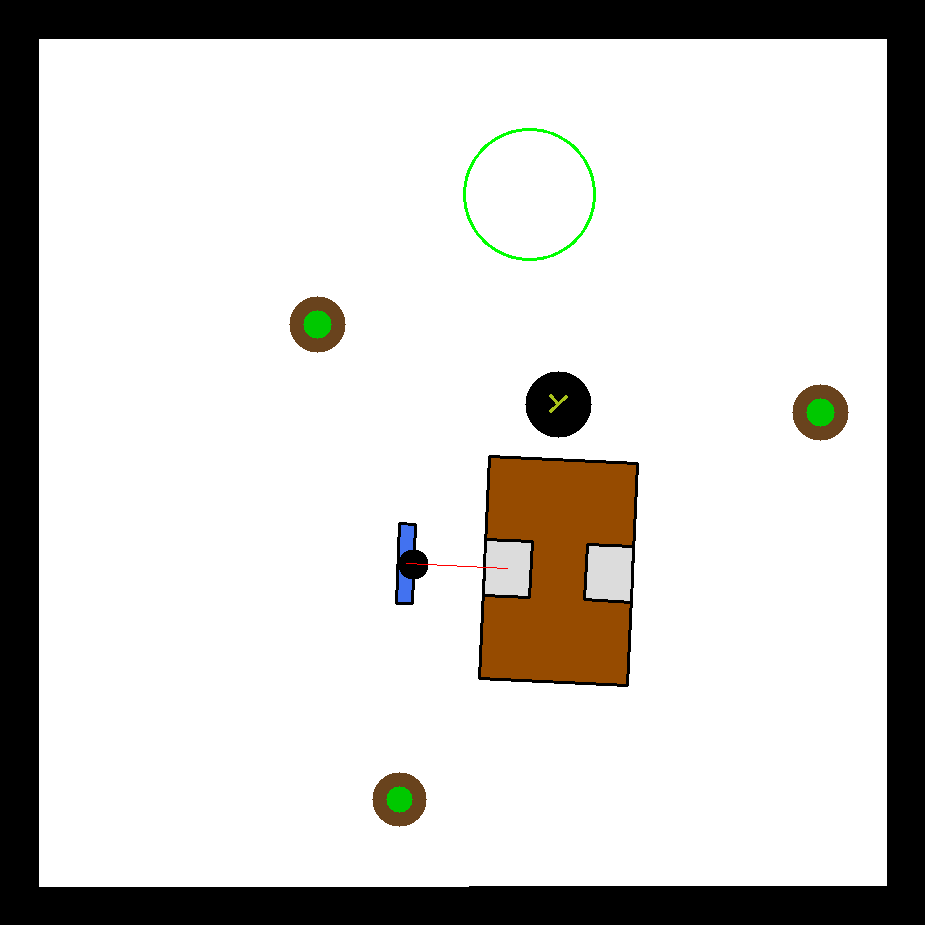}}
  \subcaptionbox{\centering Before Human-Laptop Interaction Formation}[.2\linewidth][c]{%
    \includegraphics[width=\linewidth]{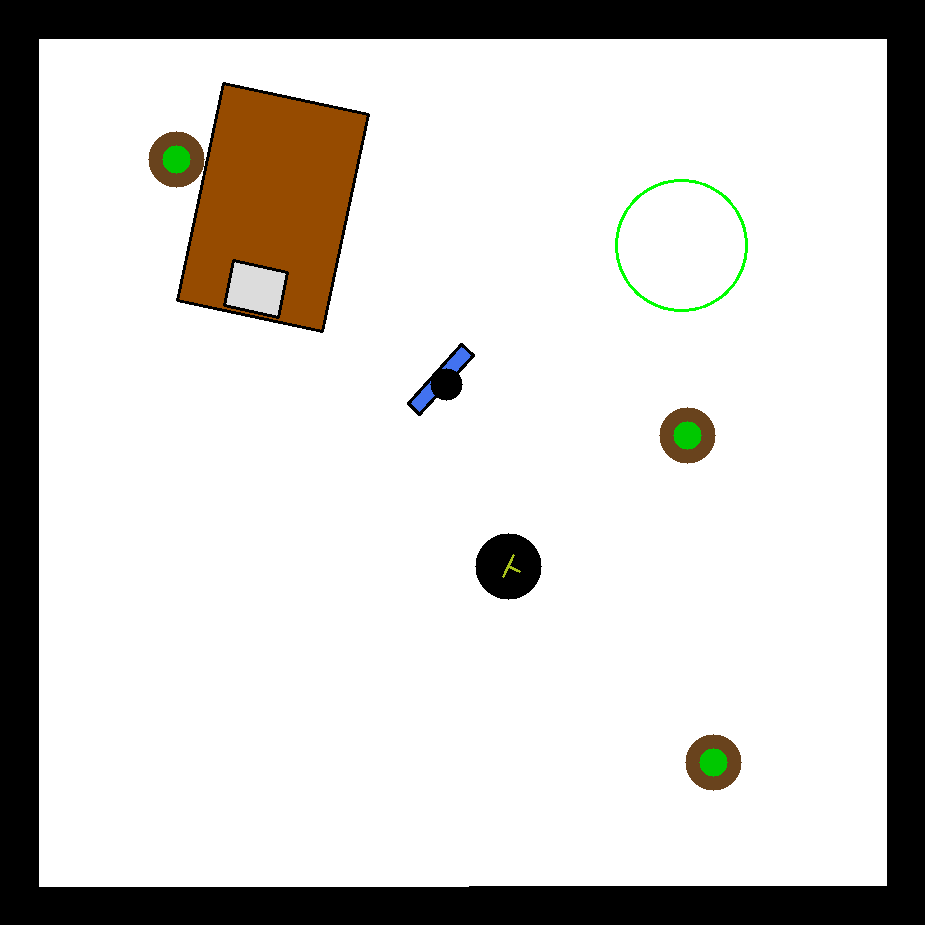}}

  \bigskip
  
  \subcaptionbox{\centering After Crowd Dispersal}[.2\linewidth][c]{%
    \includegraphics[width=\linewidth]{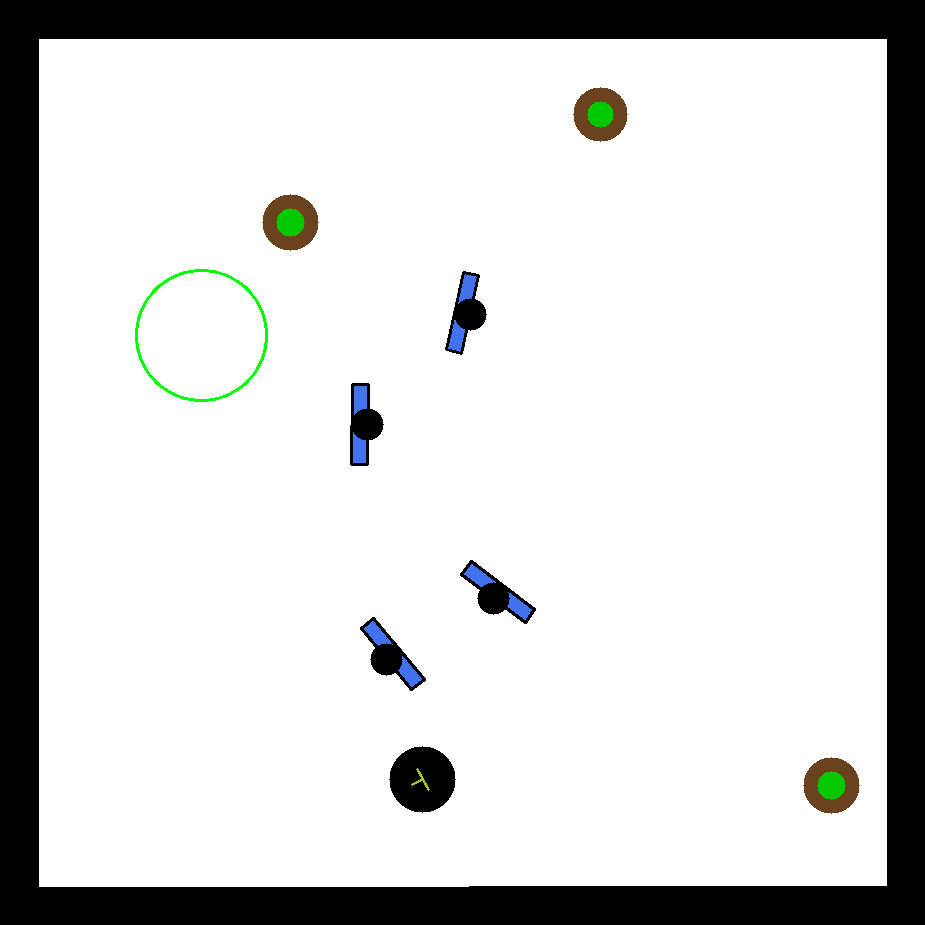}}
  \subcaptionbox{\centering After Crowd Formation}[.2\linewidth][c]{%
    \includegraphics[width=\linewidth]{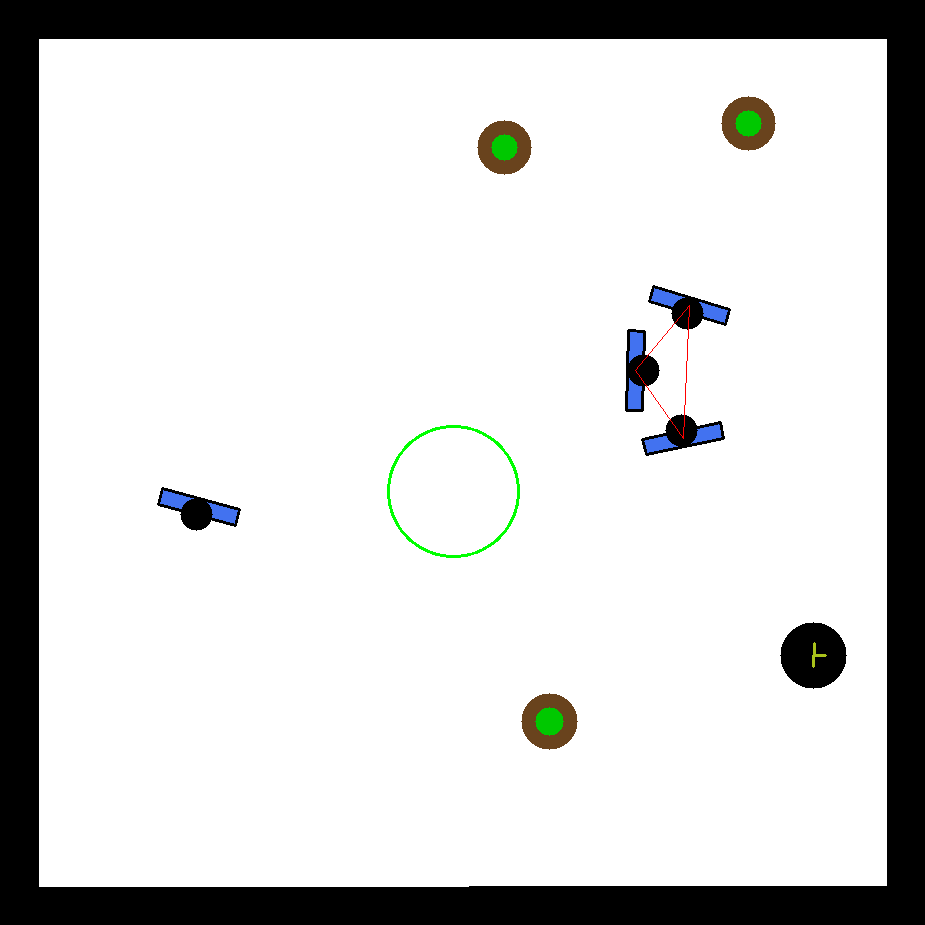}}
  \subcaptionbox{\centering After Human-Laptop Interaction Dispersal}[.2\linewidth][c]{%
    \includegraphics[width=\linewidth]{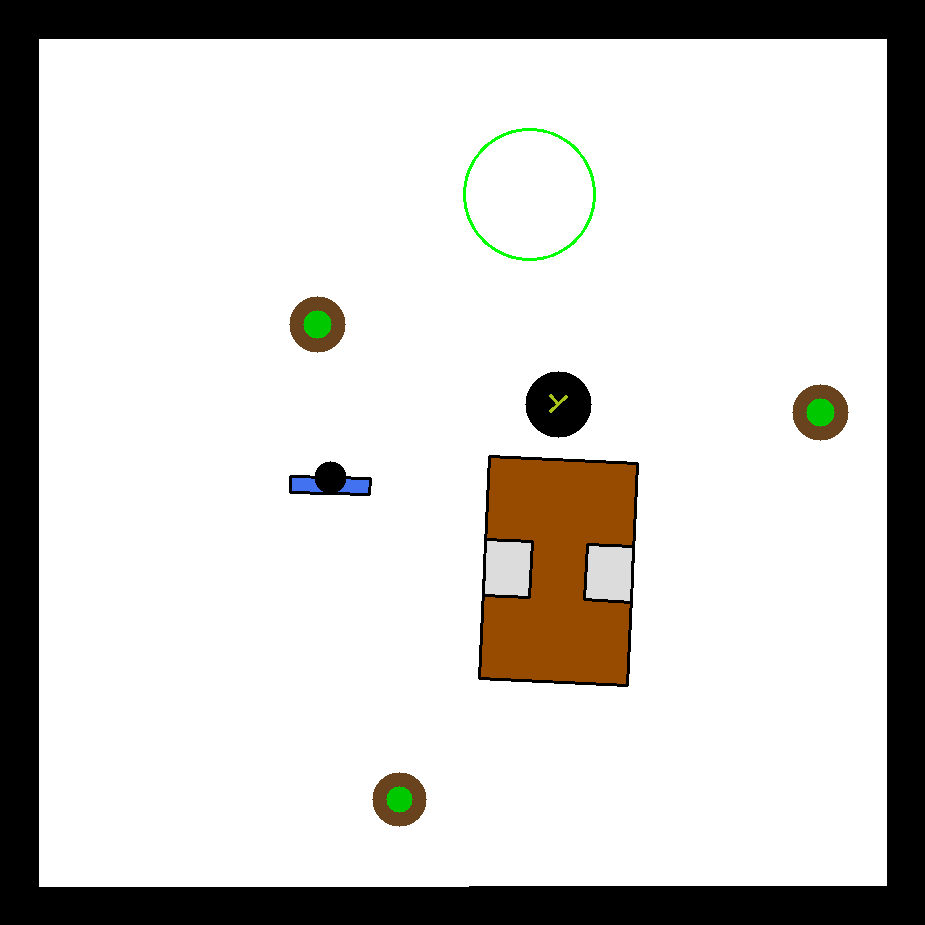}}
  \subcaptionbox{\centering After Human-Laptop Interaction Formation}[.2\linewidth][c]{%
    \includegraphics[width=\linewidth]{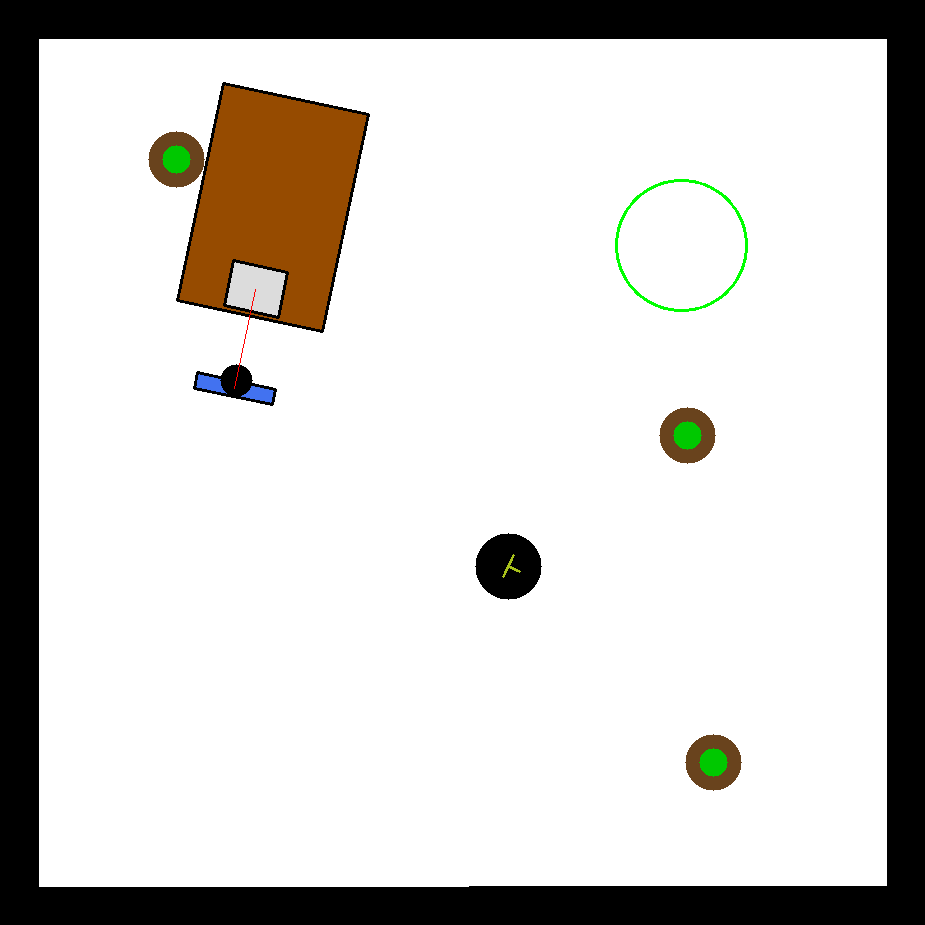}}

  \bigskip
  \subcaptionbox{\centering Corridors in squared area}[.2\linewidth][c]{%
    \includegraphics[width=\linewidth]{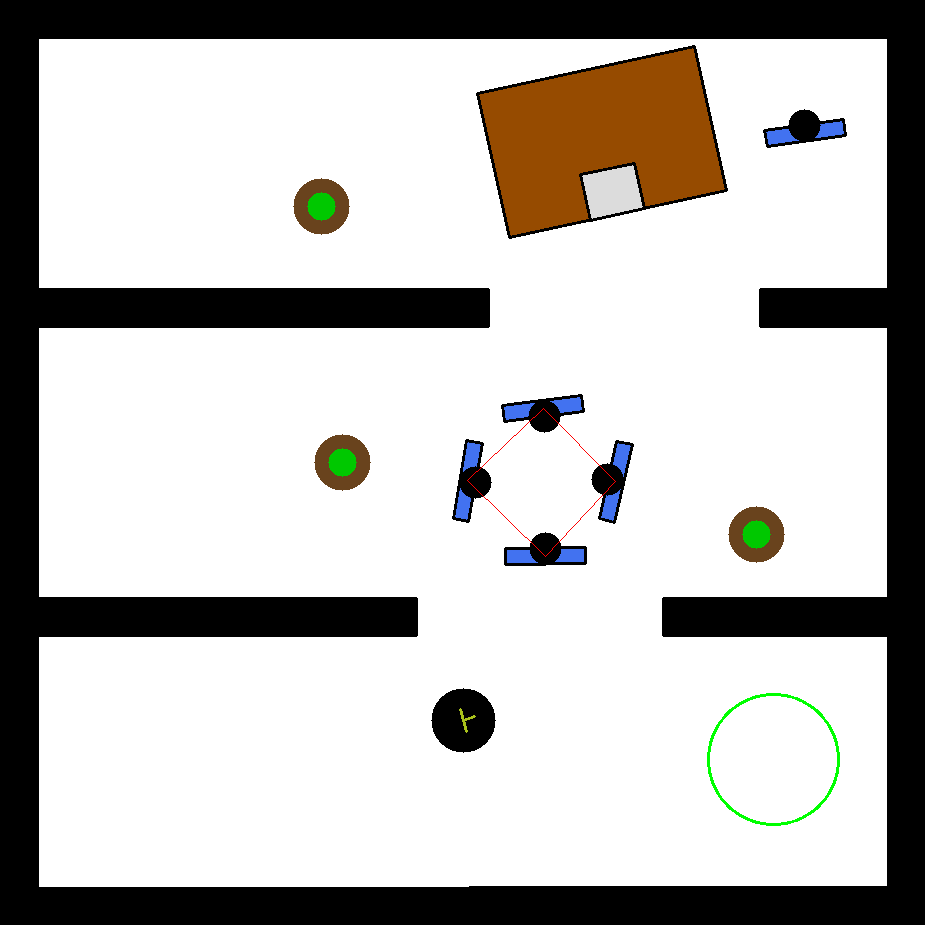}}
  \subcaptionbox{\centering L-shaped area}[.2\linewidth][c]{%
    \includegraphics[width=\linewidth]{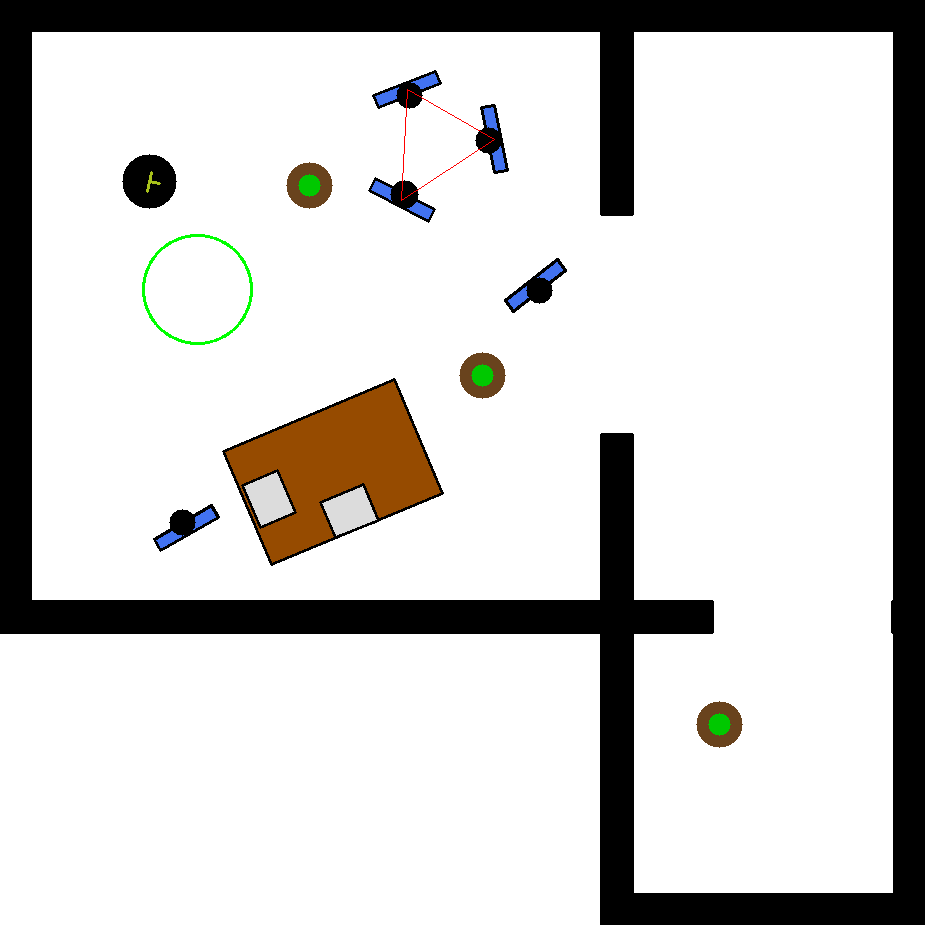}}
  \subcaptionbox{\centering Robot Limited Field of View and Sensor Range}[.2\linewidth][c]{%
    \includegraphics[width=\linewidth]{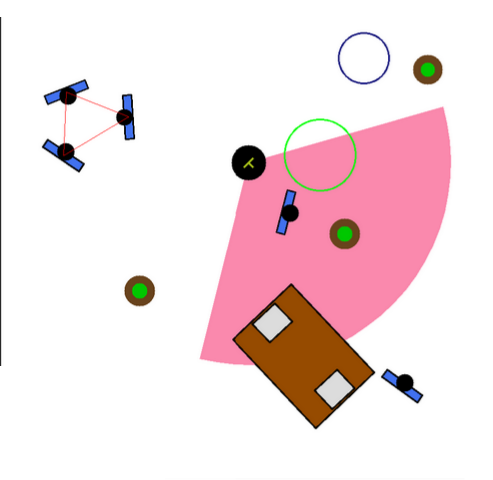}}
  \subcaptionbox{\centering Human Gaze}[.2\linewidth][c]{%
    \includegraphics[width=\linewidth]{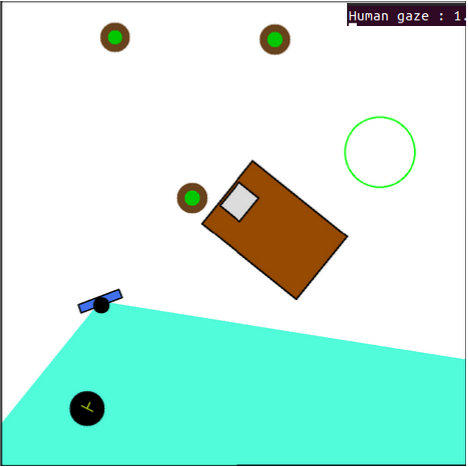}}
  
  \bigskip
  \subcaptionbox{\centering Entities' Labels}[.5\linewidth][c]{%
    \includegraphics[width=\linewidth]{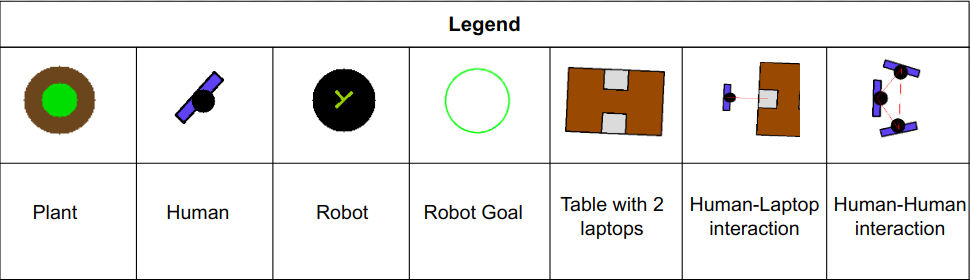}}
  \caption{Before (a-d) and after (e-h) scenario of dynamic formation and dispersal of a crowd and human-laptop interactions in the environment. Fig.i and Fig.j are square and L-shaped areas. Fig.k displays the sensor ranges of the robot and Fig.l shows the human gaze.}
  \label{Dynamic_Social_Scenes}
\end{figure*}

SocNavGym follows the standard OpenAI Gym API convention.
Installation and setup are also straightforward, either from source or via pip. 
All adjustable parameters are described in the project's documentation and can be modified via configuration files without accessing the codebase.
The project includes several sample configuration files to facilitate the use of the proposed tool.
Moreover, it provides a DuelingDQN~\cite{DuelingDQN} baseline for reference, which is described in Sec.~\ref{experiments}. 
We put forward a new open-source\footnote{\url{https://github.com/gnns4hri/SocNavGym}} gym that meets the desiderata described in Sec.~\ref{relatedworks}.
In particular, SocNavGym provides the following features:
\begin{enumerate}
    \item \label{gym_features_1} lightweight randomised scenario generation and simulation,
    \item \label{gym_features_2} support for static objects, corridors and walls to bound rooms of different shapes and sizes (Fig.~\ref{Dynamic_Social_Scenes}.i-j)),
    \item \label{gym_features_3} creation and dispersion of group formations and interactions (\textit{e.g.}, human-looking-at-laptop, group of humans) (Fig.~\ref{Dynamic_Social_Scenes} (a \& e), (b \& f), (c \& g), (d \& h)),
    \item \label{gym_features_4} support for holonomic and non-holonomic robots that can be controlled with discrete or continuous action spaces,
    \item \label{gym_features_5} integration of a learned reward function combining social and functional aspects (SNGNN-v2~\cite{SNGNN}), simple heuristic-based reward functions used in other works~\cite{chen_social_nav_2017}, and an API to create custom reward functions,
    \item \label{gym_features_6} it provides a variety of additional evaluation metrics to benchmark agents' performance (in particular the metrics suggested in the guidelines paper~\cite{francis2023guidelines}),
    \item \label{gym_features_7} it includes gaze as a human property that can be added to the observation space and allows to simulate limited Fields of View (FoV) (Fig.~\ref{Dynamic_Social_Scenes} (l)),
    \item \label{gym_features_8} it allows limiting the FoV and sensor range of the agent (Fig.~\ref{Dynamic_Social_Scenes} (k)),
    \item \label{gym_features_9} ORCA~\cite{ORCA} and social force model~\cite{SFM} policies support for humans, with the option to consider or ignore robots,
    \item \label{gym_features_10} optional and configurable noise for agents' sensory input,
    \item \label{gym_features_11} manual control of the agent to collect LfD datasets.
\end{enumerate}

\subsection{Description of Gym features}
SocNavGym is not dependent on a high-fidelity physics engine or any software framework to avoid additional overheads and delays while running simulations (feature~\ref{gym_features_1}).
Previous works like SARL~\cite{CrowdNav} and DSRNN~\cite{chen_social_nav_2017} that use the likes of the proposed simulator only work with human entities in the environment.
In our gym environment, the generated social scenarios can have different shapes and sizes across different episodic runs along with inanimate entities like plant, table, laptop and animate entities like human crowds of variable size (feature~\ref{gym_features_2}).
In realistic social settings, human crowds and human-object interactions can form and break at any point in time, and our gym environment supports such dynamics (feature~\ref{gym_features_3}).
This dynamic nature of interaction formation and dispersion is necessary to learn meaningful social context since the level of disturbance changes when an interaction undergoes a transition, that is, formation or dispersion.

\par
Regarding the steering system of the robotic agents, previous works only support holonomic agents whose learnt policy cannot be directly transferred to non-holonomic agents.
Thus we allow users to choose the agent they want to learn policies for (feature~\ref{gym_features_4}).

\par
The reward functions used in previous works are arguably designed for collision avoidance rather than discomfort avoidance.
To learn policies that truly account for social norms, we incorporated a discomfort estimator into our reward function that penalises agents for causing disturbance based on the context of the environment (feature~\ref{gym_features_5}).
We further show a use case where we train a DuelingDQN agent with the proposed reward function and one of the commonly used in the literature and benchmark its behaviour using the evaluation metrics proposed in the guidelines paper~\cite{francis2023guidelines}, see Table~\ref{tab:AgentPerformance}.
The gaze of the human can be a good source of information to make navigation decisions.
Therefore, for each human observed, SocNavGym provides a flag specifying whether the robotic agent is in their field of view (feature~\ref{gym_features_7}).
The gym also enables users to mimic the constraints of their physical hardware on robots that might have limited field of view and range (feature~\ref{gym_features_8}).
Humans in the gym can be controlled via ORCA or SFM policies (feature~\ref{gym_features_9}) and the users can choose whether the human behaviour should factor in the agent while calculating its policy.

\par
To collect offline datasets, SocNavGym provides a sample script that allows the users to manually manipulate the agent in the room to collect (state, action, next state, reward) tuples (feature \ref{gym_features_11}).

\par
The observations take the form of dictionaries containing a key per entity type, the corresponding value being a list of such types of entities, namely "goal", "humans", "objects", "walls" and "relationships".
All entities have basic properties including an identifier, 2D position coordinates, orientation, the radius of their bounding sphere, and their relative velocities -linear and angular.
The observations returned by the environment can contain information about all the entities in the environment (fully observable) or only a subset of observable entities.
This is configured by adjusting the range and field of view width of the robot, based on users' discretion (feature~\ref{gym_features_8}.
Similarly, SocNavGym can be set up so that the observations are relative to the agent or a global frame of reference.
In addition, the sensory information of specific entities can be corrupted with Gaussian noise by setting the distribution's parameters (feature~\ref{gym_features_10}).
These variations of the observations can be configured via gym wrappers. 
If SocNavGym is configured to provide it, the "relationships" section of the observation space accounts for relational information (\textit{e.g.}, humans talking, a human looking towards objects and human gaze, feature~\ref{gym_features_7}).

\subsection{Evaluation metrics}
SocNavGym provides events and the evaluation metrics suggested in~\cite{francis2023guidelines} via the \texttt{info} variable returned with each step of an episode (see Table~\ref{tab:info_dict}).
They are designed to better understand agents' performance while training (feature~\ref{gym_features_6}).
Along with these events and evaluation metrics, global metrics, corresponding to the execution of several episodes, can be obtained from the ones provided by the info variable. It also returns an adjacency matrix to identify the interacting entities in the environment at every instance of the simulation.
These global metrics are accessible via helper functions to evaluate social agents in multiple runs.

{\small
\begin{table*}[htb]
\centering
\resizebox{\textwidth}{!}{%
\begin{tabular}{| c | p{12cm} |}
\hline
 \begin{tabular}[c]{@{}c@{}} \textbf{Metric Name} \end{tabular} &
 \begin{tabular}[c]{@{}c@{}} \textbf{Description} \end{tabular} \\ 
\hline
 OUT\_OF\_MAP & True if the agent is out of the map \\ \hline
 \begin{tabular}[c]{@{}c@{}}COLLISION\_HUMAN \& COLLISION\_OBJECT \\ \& COLLISION\_WALL \end{tabular} & True if the agent has collided with a human OR object OR wall \\ \hline
 COLLISION & True if the agent has collided with any entity \\ \hline
 SUCCESS & True if the agent has reached the goal \\ \hline
 TIMEOUT& True if the episode has terminated due to max episode length \\ \hline
 FAILURE\_TO\_PROGRESS & The number of timesteps that the robot failed to reduce the distance to goal \\ \hline
 STALLED\_TIME & The number of timesteps that the robot's velocity is 0 \\ \hline
 TIME\_TO\_REACH\_GOAL & Number of time steps taken by the robot to reach its goal \\ \hline
 STL & Success weighted by time length \\ \hline
 SPL & Success weighted by path length \\ \hline
 PATH\_LENGTH & Total path length covered by the robot \\ \hline
 V\_MIN \& A\_MIN \& JERK\_MIN & Minimum velocity OR acceleration OR jerk that the robot has achieved \\ \hline
 V\_AVG \& A\_AVG \& JERK\_AVG & Average velocity OR acceleration OR jerk of the robot \\ \hline
 V\_MAX \& A\_MAX \& JERK\_MAX & Maximum velocity OR acceleration OR jerk that the robot has achieved \\ \hline
 TIME\_TO\_COLLISION & Minimum time to collision with a human at any point in time in the trajectory. \\ \hline
 MINIMUM\_DISTANCE\_TO\_HUMAN & Minimum distance to any human. \\ \hline
 PERSONAL\_SPACE\_COMPLIANCE & Percentage of steps that the robot is not within the personal space (0.45m) of any human. \\ \hline
 MINIMUM\_OBSTACLE\_DISTANCE & Minimum distance to any object. \\ \hline
 AVERAGE\_OBSTACLE\_DISTANCE & Average distance to any object. \\ \hline
 DISCOMFORT\_DSRNN & The discomfort reward according to the DSRNN reward function \\ \hline
 distance\_reward & difference in agent's distance to goal in the previous and current time steps \\ \hline
 sngnn\_reward & The SNGNN reward that is $SNGNN\_value - 1$. (~ $DISCOMFORT\_SNGNN - 1$) \\ \hline
 alive\_reward & Alive reward that the agent received (meant to be used as a penalty) \\ \hline
 closest\_human\_dist & Closest distance to a human \\ \hline
 closest\_obstacle\_dist & Closest distance to an obstacle \\ \hline
\end{tabular}%
}
\caption{Metrics in info dictionary}
\label{tab:info_dict}
\end{table*}
}

\subsection{Human trajectory generation}
Simulated pedestrian trajectories can be generated with ORCA or a social force model.
To get variable behaviours for each human, SocNavGym can be configured to use one of the two or to be chosen at random for every spawned pedestrian.
The user can also choose for the robot to be ignored when calculating the simulated humans' actions. 
On spawning, the dynamic humans and crowds are given a goal location and, on reaching the goal, if the episode is not terminated, another goal location is sampled.
SocNavGym can also be configured to create and dissolve groups, as well as human-human and human-object interactions.
For crowd dispersion, each human or sub-group gets a different goal that they have to traverse or remain in place (static).
Further, the user can change the FoV of the pedestrians to restrict its policy to only consider entities in their FoV.

\begin{figure*}[!t]
  \centering
  \subcaptionbox{\centering Snapshot of the current state of the environment}[.3\linewidth][c]{%
    \includegraphics[width=0.75\linewidth]{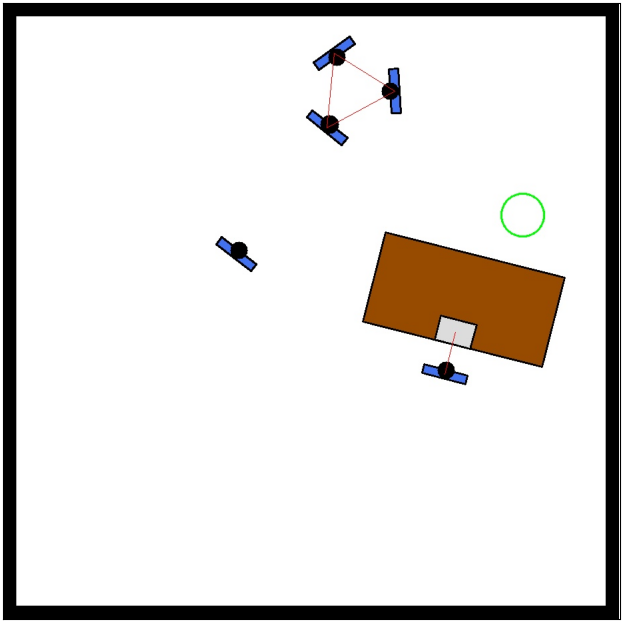}}
  \subcaptionbox{\centering SNGNNv2-discomfort values for the current state of the environment.}[.3\linewidth][c]{%
    \includegraphics[width=0.9\linewidth]{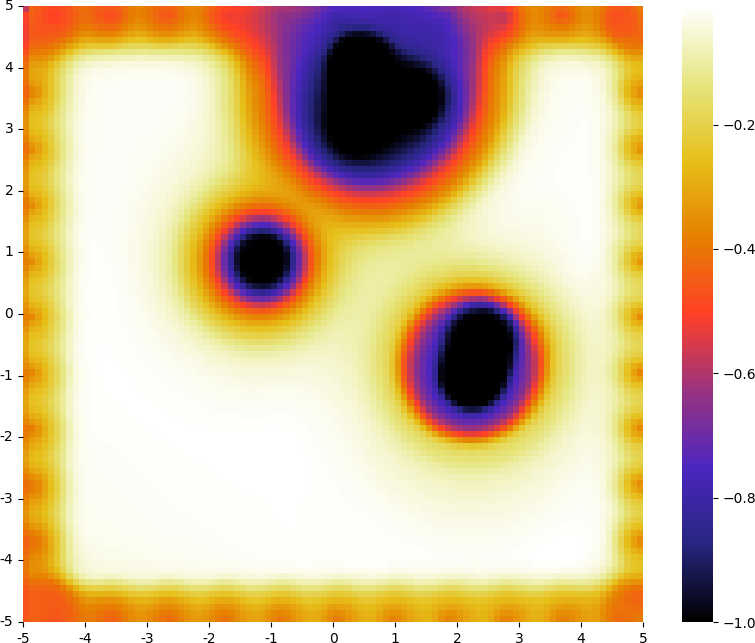}}
  \subcaptionbox{\centering DSRNN-discomfort values for the current state of the environment}[.3\linewidth][c]{%
    \includegraphics[width=0.9\linewidth]{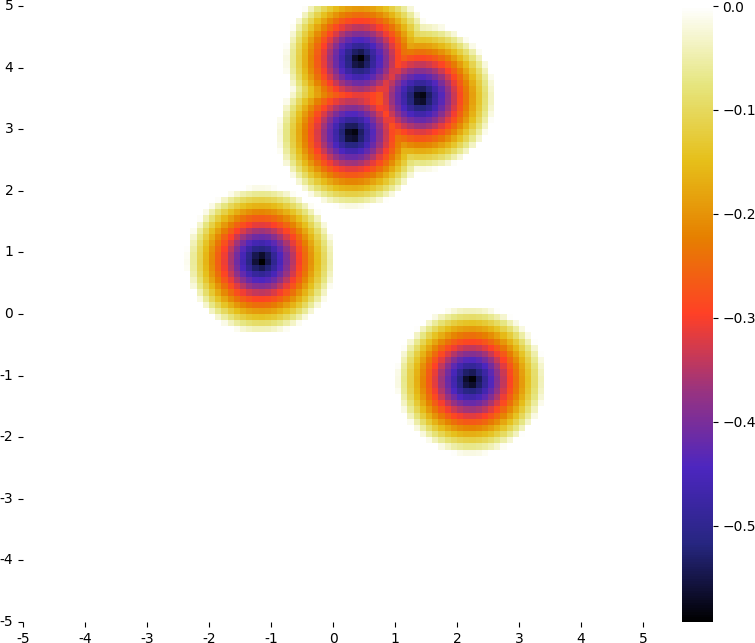}}
  \caption{Heatmaps of DSRNN-discomfort and SNGNNv2-discomfort for a social situation. Although the shape is similar (circular), the profile of the data-driven reward function is clearly non-linear. For visualisation purposes, we divided the room into an NxN grid (N=100) and placed the robot on each of these grid cells to obtain the discomfort values. We produce these values with respect to humans and interactions only so that a clean heatmap is obtained.}
  \label{fig:Heatmaps}
\end{figure*}

\begin{figure*}[!t]
  \centering
  \subcaptionbox{\centering A robot moving towards stationary pedestrians.}[.32\linewidth][c]{%
    \includegraphics[width=\linewidth]{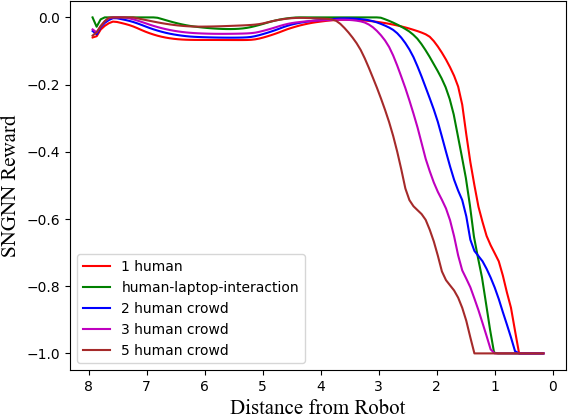}}
  \subcaptionbox{\centering Pedestrians moving towards an stationary robot.}[.32\linewidth][c]{%
    \includegraphics[width=\linewidth]{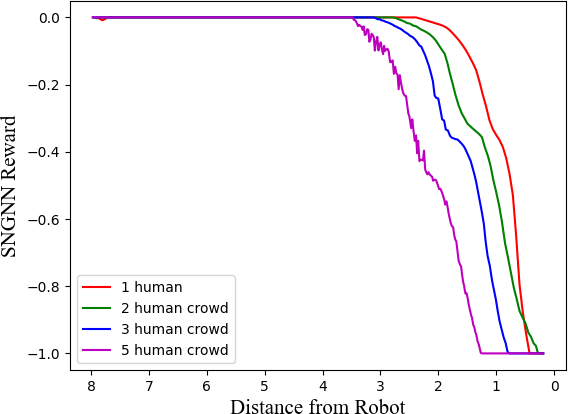}}
  \subcaptionbox{\centering Both the pedestrians and robot moving towards each other.}[.32\linewidth][c]{%
    \includegraphics[width=\linewidth]{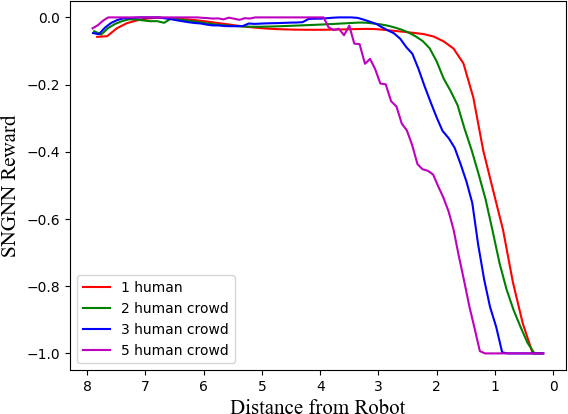}}
  \caption{SNGNN-Discomfort when the agent (a) is moving towards the stationary entities, (b) is stationary and the entities are moving towards the agent, and (c) the entities are moving towards each other. SNGNN-Discomfort considers the context as well as the presence or absence of different relations while evaluating the scenario.}
  \label{Sd}
\end{figure*}

\begin{figure*}[ht!]
  \centering
  \subcaptionbox{\centering Experiment 1}[.2\linewidth][c]{%
    \includegraphics[width=\linewidth]{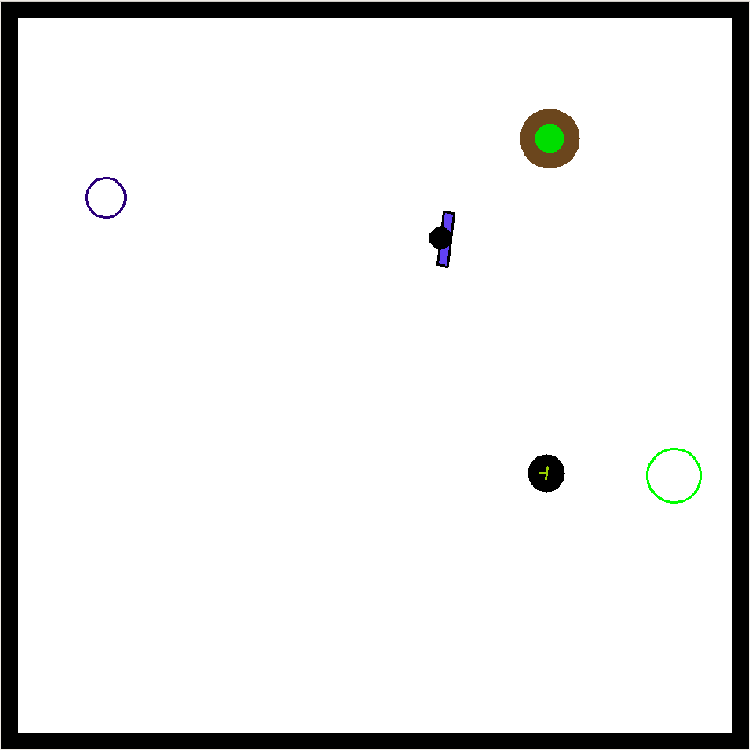}}
  \subcaptionbox{\centering Experiment 2}[.2\linewidth][c]{%
    \includegraphics[width=\linewidth]{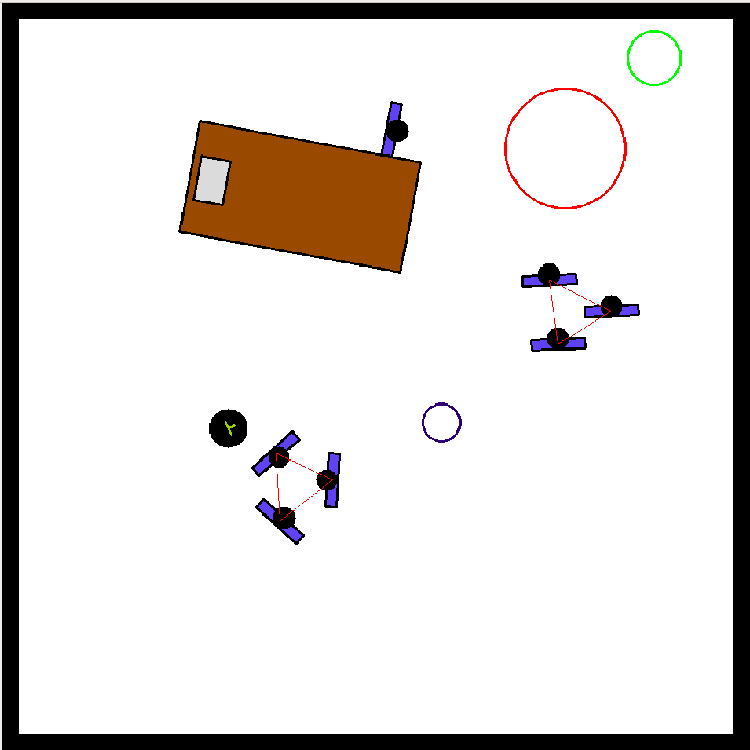}}
  \subcaptionbox{\centering Experiment 3}[.2\linewidth][c]{%
    \includegraphics[width=\linewidth]{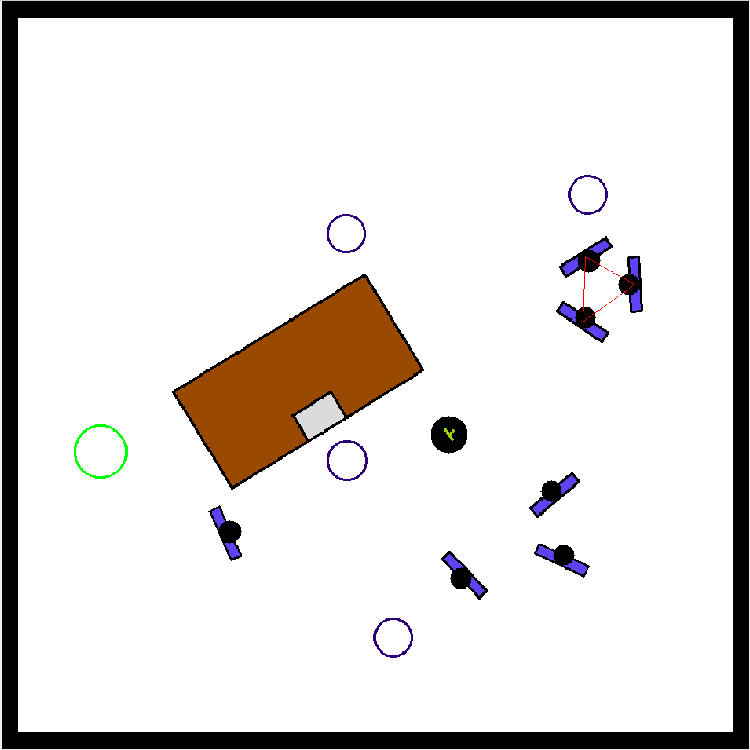}}
  \caption{Snapshot of the simulation environments used in the experiments.}
  \label{social_scenarios}
\end{figure*}

\subsection{Reward functions}
SocNavGym allows making use of the reward function used by popular DRL-based SN approaches, (\textit{e.g.}, \cite{liu2020decentralized, CrowdNav}), a piece-wise function that: a)~gives a positive reward if the robot reaches the goal; b)~provides a linear penalty if it gets too close to a human; and c)~rewards reducing the distance to the goal.
In mathematical notation:
\begin{equation}
\label{eq:reward_no_SNGNN}
\begin{split}
\begin{gathered}
    r(s_t, a_t)  = 
        \begin{cases}
            -1, \:\:\: \text{\textbf{if} } d_{min}^t \le r_{agent} + r_{entity}
            \\
            (d_{min}^t-\delta_{disc}) \cdot \alpha \cdot dt, \:\:\: \text{\textbf{if} } 0<d_{min}^t<\delta_{disc}
            \\
            1.0, \:\:\: \text{\textbf{if} } d_{goal}^t \leq \rho_{agent}
            \\
            (d_{goal}^{t-1}-d_{goal}^t) \cdot \beta, \:\:\: \text{\textbf{otherwise}}
        \end{cases}
\end{gathered}
\end{split}
\end{equation}
where $r(s_t, a_t)$ is the reward that the agent gets when performing action $a_t$ in state $s_t$ at time $t$, $d_{min}^t$ is the minimum acceptable distance between the robot and the closest human at time $t$, $r_{agent}$ and $r_{entity}$ are the radius of the agent and an arbitrary entity in the environment respectively, $\delta_{disc}$ (0.6m by default) is the threshold discomfort distance between the agent and the human, $\alpha$ is a discomfort scaling factor, $dt$ is the simulation timestep,  $\rho_{agent}$ is the goal radius, $d_{goal}^t$ is the L2 distance to the goal position at time $t$, and $\beta$ is a scaling factor for the L2 distance.
\par
The second piece of the reward function is a potential field around every entity in the room of radius $\delta_{disc}$ within which the agent is penalised as its proximity from the entity decreases. This piece is the discomfort penalty that is used as a feedback to learn social compliant behaviour.
\par

We argue that this reward function does not accurately account for human discomfort because it is agnostic to the context of the social situation. For instance, the threshold $d_{min}^t$ is chosen arbitrarily and there are many variables that the function does not consider, such as velocities, orientations, or interactions.
Another instance would be the agent getting a similar penalty when a human is approached from the back and front when, as studies have pointed out~\cite{hall1968proxemics}, the disturbance caused in both situations varies.
Thus, we propose using a reward function that accounts for users' opinions based on the overall social context.
As aforementioned, given the difficulty to model complex social preferences with hand-crafted equations, we propose the reward function in Eq.~\ref{eq:reward_SNGNN}, which replaces the above discomfort factor with an SNGNN-v2~\cite{SNGNN} based function.
Leveraging a learned discomfort metric from human feedback arguably leads to better handling of social context:
\begin{equation}
\label{eq:reward_SNGNN}
\begin{split}
\begin{gathered}
    r(s_t, a_t)  = 
        \begin{cases}
            -1, \:\:\: \text{\textbf{if} } d_{min}^t \le r_{agent} + r_{entity}
            \\
            1, \:\:\: \text{\textbf{if} } d_{goal}^t \leq \rho_{agent}
            \\
            (d_{goal}^{t-1}-d_{goal}^t) \cdot \beta + P(s_t), \:\:\: \text{\textbf{otherwise}}
        \end{cases}
\end{gathered}
\end{split}
\end{equation}
where $P(s_t)$ is the SNGNN discomfort penalty, defined as:
\begin{equation}
P(s_t) = (SNGNN(s_t)-1) \cdot \delta
\end{equation}
$SNGNN(\cdot)$ being the function implemented by the SNGNN model and $\delta$ being a scaling factor.
SNGNN scores range from 1 (no disturbance) to 0 (extreme disturbance).
Therefore, to use it as a penalty, $1$ is subtracted from its output.

\par
SNGNN-v2~\cite{SNGNN} is trained to consider social phenomena.
An example shown in Fig.~\ref{Sd} is how it accounts for the difference in the estimated discomfort in different situations given the robot-human distance.
Similarly, Fig.~\ref{fig:Heatmaps} shows heatmaps of the discomfort values used in the two reward functions.
DSRNN discomfort values are visibly homogeneous irrespective of the orientation of the human or the presence or absence of a crowd or interactions.
This illustrates that the DSRNN's reward function, although it is useful to study collision-free navigation, is arguably not the best to tackle the problem of Social Navigation.

\begin{table*}[ht!]
\centering
\begin{tabular}{|l|l|l|l|l|l|l|}
\hline
\textbf{\begin{tabular}[c]{@{}l@{}}\end{tabular}} &
  \textbf{\begin{tabular}[c]{@{}l@{}}Exp 1\\ DSRNN\end{tabular}} &
  \textbf{\begin{tabular}[c]{@{}l@{}}Exp 1\\ SNGNN\end{tabular}} &
  \textbf{\begin{tabular}[c]{@{}l@{}}Exp 2\\ DSRNN\end{tabular}} &
  \textbf{\begin{tabular}[c]{@{}l@{}}Exp 2\\ SNGNN\end{tabular}} &
  \textbf{\begin{tabular}[c]{@{}l@{}}Exp 3\\ DSRNN\end{tabular}} &
  \textbf{\begin{tabular}[c]{@{}l@{}}Exp 3\\ SNGNN\end{tabular}} \\ \hline
\textbf{Discomfort SNGNN}          & -5.495         & \textbf{-2.377}  & -17.786        & \textbf{-4.639}  & -22.376        & \textbf{-5.429}  \\ \hline
\textbf{Discomfort DSRNN}          & -0.2           & \textbf{-0.143}  & -0.481         & \textbf{-0.403}  & -1             & \textbf{-0.341}  \\ \hline
\textbf{Personal Space Compliance} & 0.979          & \textbf{0.996}   & \textbf{0.997} & 0.973            & \textbf{0.998} & 0.989            \\ \hline
\textbf{Closest Human Distance}    & \textbf{2.556} & 2.528            & 1.456          & \textbf{1.92}    & 1.376          & \textbf{1.736}   \\ \hline
\textbf{Human Collision Rate}      & 0.02           & 0.02             & \textbf{0}     & 0.06             & 0.01           & 0.01             \\ \hline
\textbf{Wall Collision Rate}       & 0.02           & \textbf{0.01}    & \textbf{0}     & 0.01             & 0.04           & \textbf{0}       \\ \hline
\textbf{Object Collision Rate}     & 0.07           & \textbf{0.04}    & \textbf{0.02}  & 0.06             & 0.12           & \textbf{0}       \\ \hline
\textbf{Collision Rate}            & 0.09           & \textbf{0.06}    & \textbf{0.02}  & 0.12             & 0.13           & \textbf{0.01}    \\ \hline
\textbf{Minimum time to collision} & 186.511        & \textbf{191.195} & 172.24         & \textbf{177.492} & 174.56         & \textbf{184.601} \\ \hline
\textbf{Closest Obstacle Distance} & 1.039          & \textbf{1.298}   & 0.806          & \textbf{0.845}   & 0.751          & \textbf{0.884}   \\ \hline
\textbf{Average Obstacle Distance} & 4.651          & \textbf{4.686}   & \textbf{4.449} & 4.394            & \textbf{4.458} & 4.426            \\ \hline
\textbf{Timeout}                   & \textbf{0.1}   & 0.26             & 0.5            & \textbf{0.08}    & 0.53           & \textbf{0.04}    \\ \hline
\textbf{Stalled Time}              & \textbf{18.61} & 23.28            & 32.25          & \textbf{13.72}   & 81.19          & \textbf{0.37}    \\ \hline
\textbf{Time to Reach Goal}        & \textbf{90.55} & 97.86            & 139.15         & \textbf{79.77}   & 162.47         & \textbf{67.14}   \\ \hline
\textbf{Failure to progress}       & \textbf{5.42}  & 12.27            & 29.06          & \textbf{2.05}    & \textbf{6.55}  & 7.84             \\ \hline
\textbf{Path Length}               & \textbf{5.558} & 6.451            & 10.44          & \textbf{4.575}   & \textbf{6.288} & 6.483            \\ \hline
\textbf{STL}                       & \textbf{0.644} & 0.607            & 0.388          & \textbf{0.746}   & 0.204          & \textbf{0.888}   \\ \hline
\textbf{SPL}                       & \textbf{0.678} & 0.624            & 0.393          & \textbf{0.75}    & 0.27           & \textbf{0.871}   \\ \hline
\textbf{Minimum velocity}          & 0.03           & 0.042            & 0.03           & 0.053            & 0.006          & 0.082            \\ \hline
\textbf{Average velocity}          & 0.08           & 0.084            & 0.082          & 0.09             & 0.052          & 0.098            \\ \hline
\textbf{Maximum velocity}          & 0.1            & 0.1              & 0.1            & 0.1              & 0.1            & 0.1              \\ \hline
\textbf{Minimum acceleration}      & 0.002          & 0.0              & 0              & 0.002            & 0              & 0                \\ \hline
\textbf{Average acceleration}      & 0.023          & 0.031            & 0.064          & 0.011            & 0.018          & 0.012            \\ \hline
\textbf{Maximum acceleration}      & 0.139          & 0.139            & 0.17           & 0.115            & 0.126          & 0.135            \\ \hline
\textbf{Minimum jerk}              & 0.002          & 0.001            & 0              & 0.004            & 0.001          & 0                \\ \hline
\textbf{Average jerk}              & 0.041          & 0.062            & 0.126          & 0.019            & 0.033          & 0.023            \\ \hline
\textbf{Maximum jerk}              & 0.214          & 0.218            & 0.303          & 0.141            & 0.224          & 0.183            \\ \hline
\end{tabular}
\caption{Performance of DuelingDQN when trained with the two reward functions. All the above metrics are averaged over 100 test episodes.}
\label{tab:AgentPerformance}
\end{table*}

\section{Use Cases}
\label{experiments}
We integrate the Dueling-DQN agent with our environment to compare the performance of SNGNN based reward function with the one popularly used in the literature that we term DSRNN reward function. The agent neural network consists of 5 linear layers with 512, 256, 128, and 64 as the hidden units using LeakyRelu as the activation function.
We used the Adam optimiser with a fixed learning rate of 0.001 and trained the agent separately with the two reward functions described above. The training lasted for 50K episodes with a maximum episode length of 200 discrete timesteps.

\subsection{Social scenarios}
The agent's observation space comprises of the position, orientation, and velocity of other entities in the agent's frame of reference.
In each experiment, the position and orientation of all  the entities' are initialised randomly, and the hyperparameters for the human's motion using ORCA or SFM are sampled from a Gaussian distribution for variable behaviour.
\par

We train a Dueling DQN agent on three social scenarios that have increasing difficulty. All experiments were carried out with a non-holonomic agent.
The following environment settings have been used to train the DRL agent:
\begin{itemize}
    \item \textit{Experiment 1}: In this experiment, we consider a 10m x 10m square room with 1 plant and 1 dynamic human (Fig~\ref{social_scenarios}.(a)).
    \item \textit{Experiment 2}: A 10m x 10m square room with 1 table, 1 laptop placed randomly on the table, a static crowd of 3 humans, a dynamic crowd of 3 humans and one human that can dynamically form or break away an ongoing interaction with the laptop on the table during the episode (Fig~\ref{social_scenarios}.(b)).
    \item \textit{Experiment 3}: A 10m x 10m square room with 1 table, 1 laptop placed randomly on the table, a static crowd of 3 humans, a dynamic crowd of 3 humans, and a dynamic human that can interact with the laptop. Humans interacting with the laptop and part of dynamic crowds are allowed to form and break their object or crowd interaction midst the episode ( Fig~\ref{social_scenarios}.(c))
\end{itemize}
\par

More complex scenarios can be generated by modifying the variables in the configuration files.

\subsection{Results}
The Dueling-DQN agent was trained using the two reward functions of Eq.~\ref{eq:reward_no_SNGNN} and Eq.~\ref{eq:reward_SNGNN} in the aforementioned social scenarios.
Table~\ref{tab:AgentPerformance} shows the performance of the trained agent across different evaluation metrics.
Specifically, two different groups of metrics are considered: social compliance metrics and navigation metrics.
Social compliance encompasses metrics such as SNGNN-discomfort (Eq.~\ref{eq:reward_SNGNN}), DSRNN-discomfort (Eq.~\ref{eq:reward_no_SNGNN}), Personal Space Compliance (PSC), closest distance to human and human collision rate.
The remainder composes the navigation metrics. Minimum, average, and maximum velocity, acceleration, and jerk give additional information about the motion primitives of the agent. 
This second group is included to help users understand the agent's performance on traversing the goal location.
As observed in Table~\ref{tab:AgentPerformance}, according to the social metrics, the agent trained with the SNGNN-v2 reward function achieves higher social compliance when compared to the agent using the DSRNN reward.
In addition, the performance of the SNGNN agent is also significantly better than the DSRNN agent in the navigation metrics in Experiments 2 \& 3, which have a more complex social situation.
This indicates that the agent can effectively learn to weigh the social context together with collision avoidance by means of this new discomfort penalty, which adds an additional benefit to the use of our tool for the creation of new human-aware robot navigation algorithms.


\section{Conclusions and future works}
In this work, we provide the HRI community with an easy-to-use social navigation environment capable to support the training of social navigation agents in different user-specified environment configurations that can vary in complexity, along with a social reward function.
The experiments evidence that considering the social reward yields better social performance with little or no damage to the Success weighted by Time Length (STL).
Although satisfactory, there are some limitations of the current work that we are working towards addressing in the near future.
These include explicitly retraining SNGNN-v2 on holonomic systems and using more sophisticated navigation policies for humans.

\section*{Acknowledgment}
Most experiments were run in the Aston EPS Machine Learning Server, funded by the EPSRC Core Equipment Fund, Grant EP/V036106/1.

\bibliographystyle{IEEEtran}
\bibliography{refs}

\end{document}